\documentclass[10pt,twocolumn,letterpaper]{article}

\usepackage{wacv}              

\usepackage{graphicx}
\usepackage{amsmath}
\usepackage{amssymb}
\usepackage{booktabs}

\usepackage{multirow}
\usepackage{algpseudocode}
\usepackage[ruled,vlined]{algorithm2e}

\usepackage[pagebackref,breaklinks,colorlinks]{hyperref}

\usepackage[capitalize]{cleveref}
\crefname{section}{Sec.}{Secs.}
\Crefname{section}{Section}{Sections}
\Crefname{table}{Table}{Tables}
\crefname{table}{Tab.}{Tabs.}

\def\wacvPaperID{406} 
\def\confName{WACV}
\def\confYear{2025}

\begin{document}

\title{Make-A-Texture: Fast Shape-Aware Texture Generation in 3 Seconds}

\author{Xiaoyu Xiang, Liat Sless Gorelik, Yuchen Fan, Omri Armstrong, Forrest Iandola,\\ Yilei Li, Ita Lifshitz, Rakesh Ranjan\\
Meta \\
{\tt\small \{xiangxiaoyu, lsg, ycfan, omriarm, fni, yileil, italif, rakeshr\}@meta.com}
}

\twocolumn[{%
\renewcommand\twocolumn[1][]{#1}%
\maketitle
\vspace{-6mm}
\centering
\captionsetup{type=figure}
\vspace{-2mm}
\includegraphics[width=\textwidth]{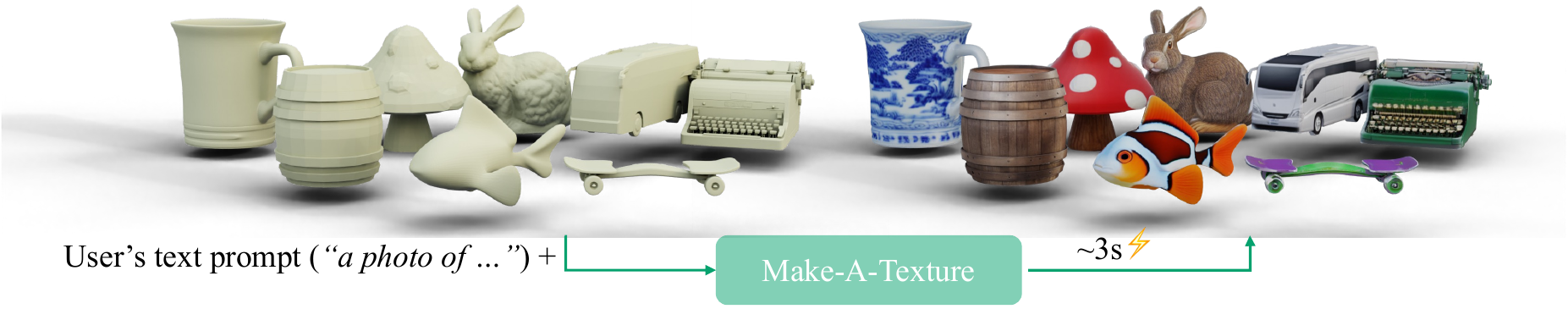}
\caption{Given a textureless 3D mesh and a text prompt, \textit{Make-A-Texture} efficiently synthesizes high-quality textures with 3 seconds.}
\label{fig:teaser}
\vspace{5mm}
}]    
    

\begin{abstract}
We present \textbf{Make-A-Texture}, a new framework that efficiently synthesizes high-resolution texture maps from textual prompts for given 3D geometries. Our approach progressively generates textures that are consistent across multiple viewpoints with a depth-aware inpainting diffusion model, in an optimized sequence of viewpoints determined by an automatic view selection algorithm. 
A significant feature of our method is its remarkable efficiency, achieving a full texture generation within an end-to-end runtime of just 3.07 seconds on a single NVIDIA H100 GPU, significantly outperforming existing methods. Such an acceleration is achieved by optimizations in the diffusion model and a specialized backprojection method. Moreover, our method reduces the artifacts in the backprojection phase, by selectively masking out non-frontal faces, and internal faces of open-surfaced objects. 
Experimental results demonstrate that Make-A-Texture matches or exceeds the quality of other state-of-the-art methods. Our work significantly improves the applicability and practicality of texture generation models for real-world 3D content creation, including interactive creation and text-guided texture editing.

\end{abstract}
\section{Introduction}
\label{sec:intro}

In games, movies, and most digital 3D experiences, objects are typically represented as meshes (geometry) and textures (surface painting). 
Considering the large number of objects required per game or movie, manually crafting and painting such a large catalog of objects can be labor-intensive and tedious. 
Recent advances in AI-drive texture synthesis works, such as SyncMVD~\cite{liu2023text}, Paint3D~\cite{zeng2023paint3dpaint3dlightingless} and Meshy~\cite{meshy}, have begun to empower artists with tools that can automatically texture arbitrary category of 3D object based on a text prompt. 
While promising, these methods generally take more than one minute to generate a full texture, which hampers the ability of artists to interactively experiment with different prompts and creative ideas.
Our goal is to develop an extremely fast texture synthesis method that can deliver results in seconds rather than minutes, with comparable quality to the state-of-the-art techniques.
We believe this will open up new possibilities for artists to quickly create high-quality libraries of themed assets, and enhance the interactive experience for gamers, allowing them to dynamically modify their virtual worlds.

Towards this goal, we introduce \textbf{Make-A-Texture}, a novel texture synthesis framework designed to address the challenges of texturing arbitrary 3D geometries efficiently and effectively with the text-to-image diffusion models. 
Our approach starts with an awkwardly simple yet effective baseline, incorporating multiple lightweight techniques for reducing artifacts and improving quality.
To transform textual prompts into high-quality, multiview-consistent textures that align with the underlying geometry, we utilize a depth-aware generative model~\cite{bensadoun2024meta3dtexturegenfast} with inpainting and depth conditioning to generate textures in a progressive manner.
Unlike previous methods that typically have a generation stage followed by a refinement stage across 10 to 30 different views of the 3D object, 
our method optimizes this process by automatically identifying just 6 optimal viewpoints for the generation stage. 
Further, to avoid baking unwanted highlights and shadows into the textures, previous methods use Voronoi filtering~\cite{liu2023text} or trimaps~\cite{chen2023text2tex, richardson2023texture}. In contrast, we implement a straightforward thresholding onto the normal map to filter out non-frontal faces.
Additionally, we propose a simpler and more efficient method for backprojecting from 2D pixels to UV coordinates of 3D meshes.
These techniques, implemented efficiently on a single GPU, yield a texture generation model with quality similar to the state-of-the-art method~\cite{liu2023text} while being 17$\times$ faster.




\textbf{Make-A-Texture} has the following advantages compared to previous works:

\noindent \textbf{Multiview Consistency}:
by generating the front and back views together, our work enforces better global consistency.
While other works tweaked depth-to-image generation model with masked generation methods, our method dynamically control the balances between depth and inpainting controls at each stage, which brings better local content consistency.

\noindent \textbf{Generality}: our method can not only generate textures for a wide range of mesh categories, but also effectively handle various mesh types, including challenging non-watertight meshes. Such high adaptability makes our method a practical solution for diverse 3D modeling applications.

\noindent \textbf{Speed}: we propose a fast backprojection method with coordinate mapping and bilinear splatting, which has the lowest latency compared to all known methods.
Our texture generation (4.56 seconds on NVIDIA A100; 3.07 seconds on H100) is significantly faster than previous methods like SyncMVD~\cite{liu2023text} (81 seconds on A100) and Meta 3D TextureGen~\cite{bensadoun2024meta} (19 seconds on H100).

\section{Related work}
\subsection{Diffusion-based Image Generation}
Recent advancements in image generation have been significantly driven by diffusion models~\cite{ho2020denoisingdiffusionprobabilisticmodels}, which have demonstrated remarkable capabilities in generating high-fidelity and diverse images. Pixel-space diffusion models such as~\cite{nichol2022glidephotorealisticimagegeneration},~\cite{ramesh2022hierarchicaltextconditionalimagegeneration} ~\cite{saharia2022photorealistictexttoimagediffusionmodels} produce photorealistic images but are computationally demanding.
To address these computational challenges, Latent-Diffusion models~\cite{rombach2022highresolutionimagesynthesislatent} apply the diffusion process in a latent space, enhancing efficiency but still keeping a high fidelity level. Other works like ControlNet\cite{zhang2023adding} and T2I-Adapter~\cite{mou2023t2iadapterlearningadaptersdig} have expanded the conditioning capabilities of diffusion models, incorporating signals like depth, edges and masking. For a thorough overview of diffusion models, we recommend the following comprehensive survey~\cite{yang2024diffusionmodelscomprehensivesurvey}.
In our work, we utilize a depth-aware inpainting model\cite{bensadoun2024meta3dtexturegenfast}, conditioned on depth maps and inpainting masks, for our text-to-texture generation process.

\subsection{Texture Generation and Editing}
In our study, we employ pretrained 2D diffusion models for text-based texture generation and editing. Alternative non-diffusion approaches include GAN-based methods such as Texturify~\cite{siddiqui2022texturifygeneratingtextures3d} and Mesh2Tex~\cite{bokhovkin2023mesh2texgeneratingmeshtextures}. Techniques like CLIP-Mesh~\cite{Mohammad_Khalid_2022} and Text2Mesh~\cite{michel2021text2meshtextdrivenneuralstylization} utilize CLIP~\cite{radford2021learningtransferablevisualmodels} to guide optimization processes based on textual descriptions. While Latent-Paint~\cite{metzer2022latentnerfshapeguidedgeneration3d}, Fantasia3D~\cite{chen2023fantasia3ddisentanglinggeometryappearance}, Paint-it~\cite{youwang2024paintittexttotexturesynthesisdeep} and TextureDreamer\cite{yeh2024texturedreamer} also integrate pre-trained diffusion models, they apply these within SDS~\cite{poole2022dreamfusiontextto3dusing2d}-based optimizations.
Recent advancements in texture generation using diffusion models are notable. TexDreamer~\cite{liu2024texdreamerzeroshothighfidelity3d} and Geometry Aware Texturing~\cite{cheskidova2023geometry} directly employ diffusion processes in UV space, achieving rapid texture generation tailored for specific domains. Point-UV Diffusion~\cite{yu2023texturegeneration3dmeshes} introduces a two-stage diffusion process, initially applying diffusion to a point cloud and subsequently refining the texture via UV diffusion. However, this method lacks generality and necessitates distinct model training for each category.
Several approaches, such as TEXTure~\cite{richardson2023texturetextguidedtexturing3d}, Text2Tex~\cite{chen2023text2textextdriventexturesynthesis}, and InTex~\cite{tang2024intex}, utilize depth-conditioned diffusion models. These methods involve an iterative process where each iteration generates a new single-view image of the object, which is then projected onto UV space for inpainting, considering previously seen views. Paint3D~\cite{zeng2023paint3dpaint3dlightingless} extends this approach by incorporating a stage to eliminate lighting artifacts.
While these methods share similarities with our approach, they generally exhibit slower processing times and produce textures with less global consistency. To enhance view consistency, SyncMVD~\cite{liu2023textguidedtexturingsynchronizedmultiview} synchronizes information across different views during the denoising process, yet it remains too slow for interactive applications. Other techniques, such as FlashTex~\cite{deng2024flashtexfastrelightablemesh}, leverage expensive 3D data. FlashTex generates four views based on text prompts with fixed lighting, followed by a texture optimization phase using SDS loss. Similarly, TextureGen~\cite{bensadoun2024meta3dtexturegenfast} produces four images from different views in its initial stage, but its second stage involves projecting these images onto UV space and employing a UV-space inpainting model conditioned on normal and position maps to finalize the texture.
In text-guided texture generation models, global texture modifications are achieved by altering the text prompts. TEXTure\cite{richardson2023texturetextguidedtexturing3d} further supports localized edits. This is achieved by incorporating a mask during the iterative process, which is then combined with a new guiding text to direct specific changes. In our study, we extend these capabilities by demonstrating how our framework can also be employed for text-guided texture enhancement, as detailed in Section~\ref{subsec: tex_enhance}.

\begin{figure*}[tbp]
	\centering
	\includegraphics[width=\linewidth]{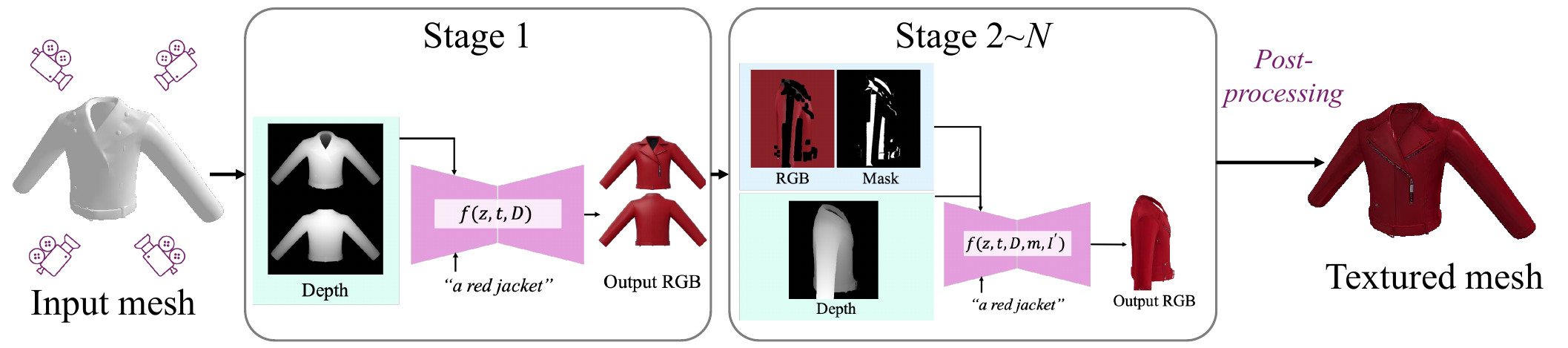}
	\caption{\textbf{Method overview}. The texture is generated iteratively from different viewpoints using a pretrained diffusion model. At the 1st stage, we generate the front and back view together for better global consistency. In following stages, the output RGB is conditioned on both geometry and the existing textures via inpainting. The generated images are backprojected to the mesh surface to for the next stage.}
	\label{fig:pipeline}
\end{figure*}

\section{Method}

Given an untextured 3D geometry $M$, we aim to generate a texture that faithfully aligns with every detail of the model, while also adhering to the provided textual description. To achieve this goal, we employ 2D text-to-image diffusion models to generate images that are consistent across multiviews (Sec.~\ref{subsec: geo-aware-mv-gen}) in a progressive manner (Sec.~\ref{subsec: pro_gen}). We then employ two meticulous filtering methods in the screen space before projecting back to the UV space to minimize artifacts and improve texture quality (Sec.~\ref{subsec: filtering}). We further post-process the textures by filling the missing areas in UV space with inpainting. Thus, all visible areas are guaranteed to have textures. Finally, we apply various acceleration and optimization techniques to reach a fast runtime of approximately 3s (Sec.~\ref{subsec:imp_speed}). 

\subsection{Geometry-Aware Multiview Generation}
\label{subsec: geo-aware-mv-gen}
To adapt a 2D generation model for 3D objects, we address two primary challenges: (1) establishing geometry-texture correspondence, and (2) ensuring multi-view consistency. We use rendered depth as a bridge between the 3D geometry and the 2D image space, and leverage a depth-aware text-to-image generation model $f$ to ensure that the generated texture are faithfully aligned with the underlying shapes. 

Given a set of camera poses $\mathcal{C}=\{C_i\}_{i=1}^{n}$ for $M$, we render the z-buffer depth $D_i$ for each view $C_i$, and sample the appearance image $I_i$ for view $C_i$ conditioned on $D_i$ and the text prompt $t_i$ using the text-to-image model.


\noindent \textbf{Global Consistency.} 
Directly conducting view-by-view image synthesis cannot guarantee consistency among the generated views due to the lack of constraints or cues to regulate the independent sampling process. 
To solve this problem, we propose to first synthesize the front and back views simultaneously, thereby implicitly enforcing global consistency. 

Given the camera poses $C_{front}$ and $C_{back}$ corresponding to the front and back views, we render the depth map $D_{front}$ and $D_{back}$ accordingly, and concatenate them into grids:

\begin{equation}
    D_0 = (D_{front} | D_{back})
\end{equation}

This depth map grid $D_0$ is inputted into a depth-conditioned image generation model, along with a textural prompt $t_0$ specifying the "front and back" of the object. The generated image grid $I_0$ can be formulated as:
\begin{equation}
    I_0 = f(z, t_0, D_0),
\end{equation}
where $z$ is a 2D Gaussian noise map. Since the text prompt $t_0$ explicitly describes the front and back views of the same object, and the dual-view generation process inherently synchronizes the textural attributes, the generated image grid $I_0$ maintains highly consistent visual appearance between different camera views.

\noindent \textbf{Cross-view Consistency.} 
This grid image generation strategy is effective for non-overlapping views. However, it lacks correspondence information from the geometry, which is crucial for overlapping views. To address this, we incoporate inpainting into the depth-to-image generation process for subsequent views, ensuring the newly generated textures can keep the same color and styles as the front and back views.


When the camera is rotated to a new position, the rendered RGB will display both previously generated textures and unpainted area. We utilize an inpainting mask to indicate where to generate the new textures.
During the texture generation process, we maintain a binary mask $T^B$ in the UV space, which represents whether the texture has been generated in the UV map. 
For a new view $C_i$, we render $T^B_i$ using the same rasterization pipeline as the RGB image, yielding a projected binary mask $m_i$. Along with the rendered depth map $D_i$ and the initial RGB image $I_i'$, the RGB image $I_i$ is generated through:

\begin{equation}
    I_i = f(z, t_i, D_i, m_i, I_i'),
\end{equation}
where $z$ denotes the 2D Gaussian noise, $t_i$ is the the prompt corresponding to the current view, $D_i$ is the rendered depth map.

For implementation, we employ a depth-aware generation model~\cite{bensadoun2024meta3dtexturegenfast} conditioned on depth maps and the inpainting mask as our text-to-image generator $f$. To acquire high-quality output that faithfully follow the geometry and are consistent with the existing textures, we dynamically adjust the control weights for depth ($w_D$)
and inpainting ($w_I$) for different views. Users can also change the weights to make the outputs better match the input prompt. We also append the view information to the text prompt (\textit{e.g.} "a side view of ...") according to the camera pose to alleviate the Janus problem.

\subsection{Progressive Generation}
\label{subsec: pro_gen}
For view $C_i$, the depth-aware inpainting output heavily depends on the textures generated from previous views $C_0, \ldots, C_{i-1}$. Hence, the sequence of camera views greatly influences the output consistency and quality. 


\noindent \textbf{Coarse-to-Fine Generation with Automatic View Selection.}
To solve the above problem, we develop an automatic view selection algorithm employing a coarse-to-fine strategy. This algorithm prioritizes views based on the number of untextured pixels, selecting the view with the largest untextured region first and progressively moving to smaller regions. This greedy approach generates textures iteratively from larger to smaller areas, maximizing the texture coverage in training views.

The primary objective of this algorithm is to identify and texture the view with the least amount of existing texture at each stage. Alg.~\ref{alg:1_auto_view_selection} outlines the steps of this automatic view selection process: 
starting with the UV texture $T$ with painted front and back views, there are $N$ candidate views. As described in Sec.~\ref{subsec: geo-aware-mv-gen}, the binary mask $T^B$ marks textured pixels as 1, and untextured pixels as 0. Using $T^B$, the algorithm selects the view the highest number of untextured pixels for subsequent texturing, updating $T^B$ accordingly. This process is repeated iteratively until most pixels are textured, or a predefined number of iterations is reached.

\begin{algorithm}[ht]
\SetAlgoLined
\textbf{Input:} mesh $M$, a binary UV texture mask $T^B$ marking the textured pixels of $T$, and $N$ candidate views $\{C_i\}_{i=1}^{N}$\;
 \For{number of iterations}{
 For each view $i$, render the binary mask $I^B_i$ from $T^B$: $\{I_i^B\}_{i=1}^N$=$\{Render(T^B, C_i)\}_{i=1}^N$ \;
 Select the least textured view $C_j$: $j \leftarrow \arg\min_{i=1}^{N} \sum I_i^B$ \;
 Generate the appearance image $I_j$ \;
 Backproject $I_j$, update $T^B$;
 }
 \caption{Automatic View Selection}
 \label{alg:1_auto_view_selection}
\end{algorithm}

This approach prioritizes filling large, more visible areas before addressing to smaller and more occluded regions. In practice, the candidate camera views are sampled on a sphere with radius $r$, where azimuth and elevation are determined based on Fibonacci lattice for given $N$, ensuring nearly uniform sampling across the sphere.

After generating RGB image $I_i$ for each view, we update the texture $T$ by backprojecting $I_i$ to the UV space using a gradient-based optimization for the differentiable renderer $\mathcal{R}$. More details of the backprojection algorithm will be discussed in Sec.~\ref{subsec:imp_speed}. 

\subsection{Filtering}
\label{subsec: filtering}

The creation of high-quality textures requires to correctly project the high-quality pixels to the texture map without introducing artifacts. 
However, the generated images often contain highlights and shadows that, if not handled correctly, can be mistakenly baked into the texture, leading to color shifts during inpainting in subsequent views.

\begin{figure}
	\centering
	\includegraphics[width=\linewidth]{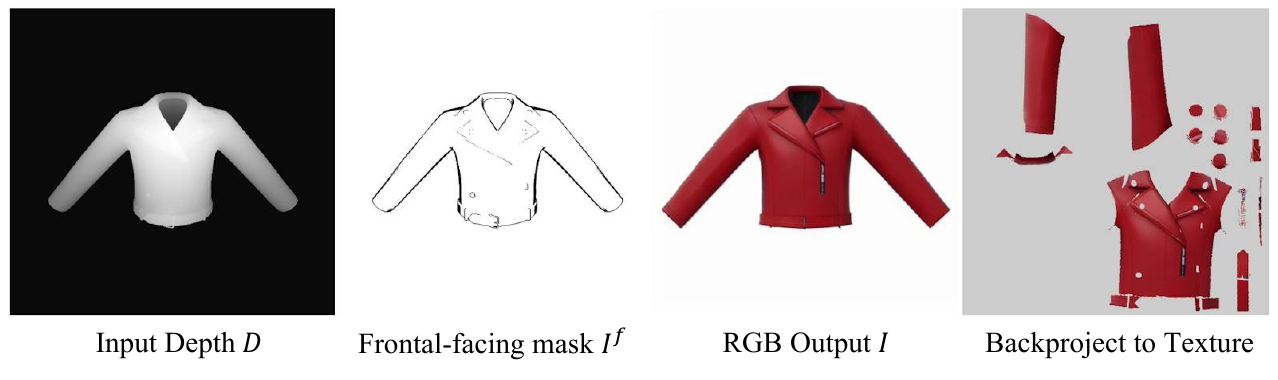}
	\caption{Filtering out non-frontal-facing regions. From the input depth, we can derive normals and a binary frontal-facing mask by thresholding. This mask guides the backprojection process. Thus, areas like the side or seam of the jacket are not mapped to the output texture, as shown in the rightmost image.}
	\label{fig:filtering}
\end{figure}

\noindent \textbf{Handling Highlights and Shadows.}  To mitigate this issue, we refine the backprojection process: it should retain pixels with high confidence while rejecting those with lower confidence or causing artifacts. Therefore, we propose to filter out the non-frontal facing views with normals derived from the depth, as illustrated in Fig.~\ref{fig:filtering}. We create a binary mask $I^f$ by thresholding the $z$ channel of the normal map:

\begin{equation}
    I^{f} = \begin{cases}
      1, & \text{if } normal_z < threshold \\
      0, & \text{otherwise}
    \end{cases}  
\end{equation}

During backprojection, the masked pixels in $I^f$ are not mapped to the UV texture. This filtering ensures that highlights and shadows in these regions minimally influence the final texture output.

\begin{figure}[tbp]
    \centering
    \begin{subfigure}[b]{0.47\linewidth}
        \includegraphics[width=\textwidth]{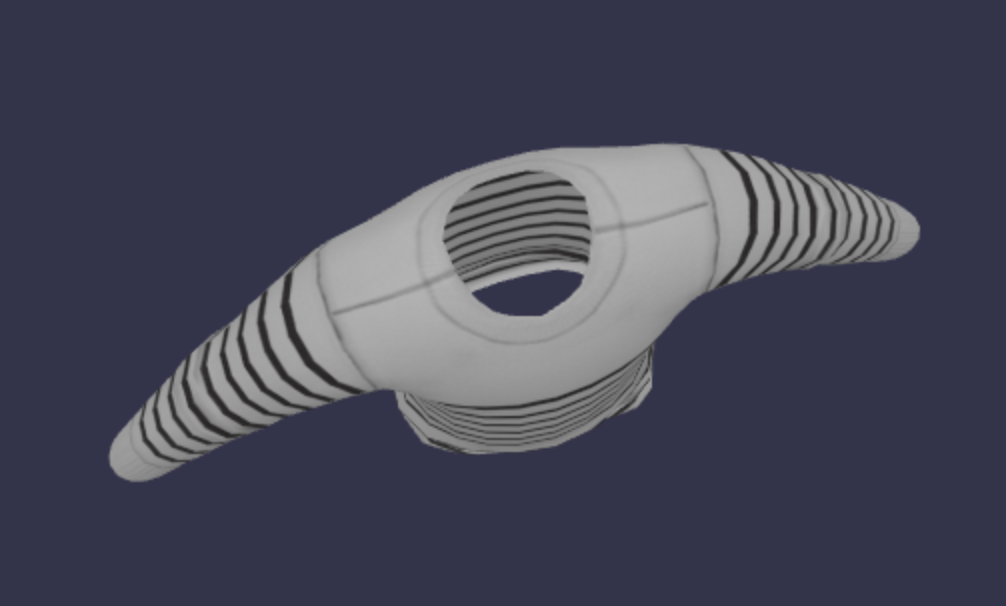}
        \caption{Mesh with open surfaces}
        \label{fig:uncapped_asset}
    \end{subfigure}
    \begin{subfigure}[b]{0.48\linewidth}
        \includegraphics[width=\textwidth]{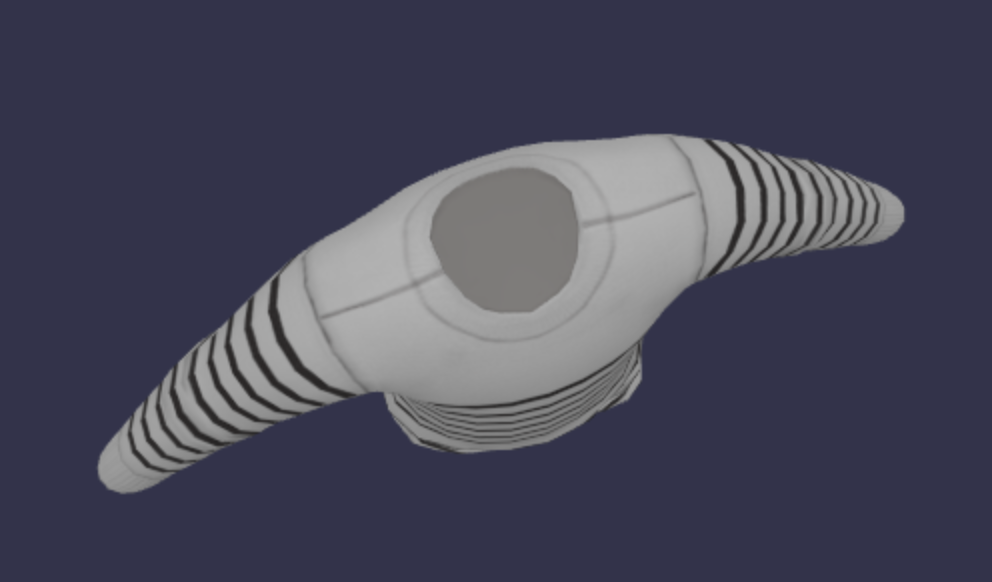}
        \caption{Mesh with closed surfaces}
        \label{fig:capped_asset}
    \end{subfigure}
    \caption{Example of different shape representations.}
    \label{fig:example_assets}
\end{figure}

\noindent \textbf{Handling open surfaced assets.} Open-surface meshes are common, \textit{e.g.} garments, where neck and sleeve holes enable draping and layering of the models on top of avatars. These meshes are not watertight (see Fig.~\ref{fig:uncapped_asset}). When dealing with shapes modeled with open surfaces, directly applying depth-conditioning and backprojection in such cases can cause artifacts: textures generated for the internal faces may be projected to external faces.

\begin{figure}[tbp]
    \centering
    \begin{subfigure}[b]{0.22\linewidth}
        \includegraphics[width=\linewidth]{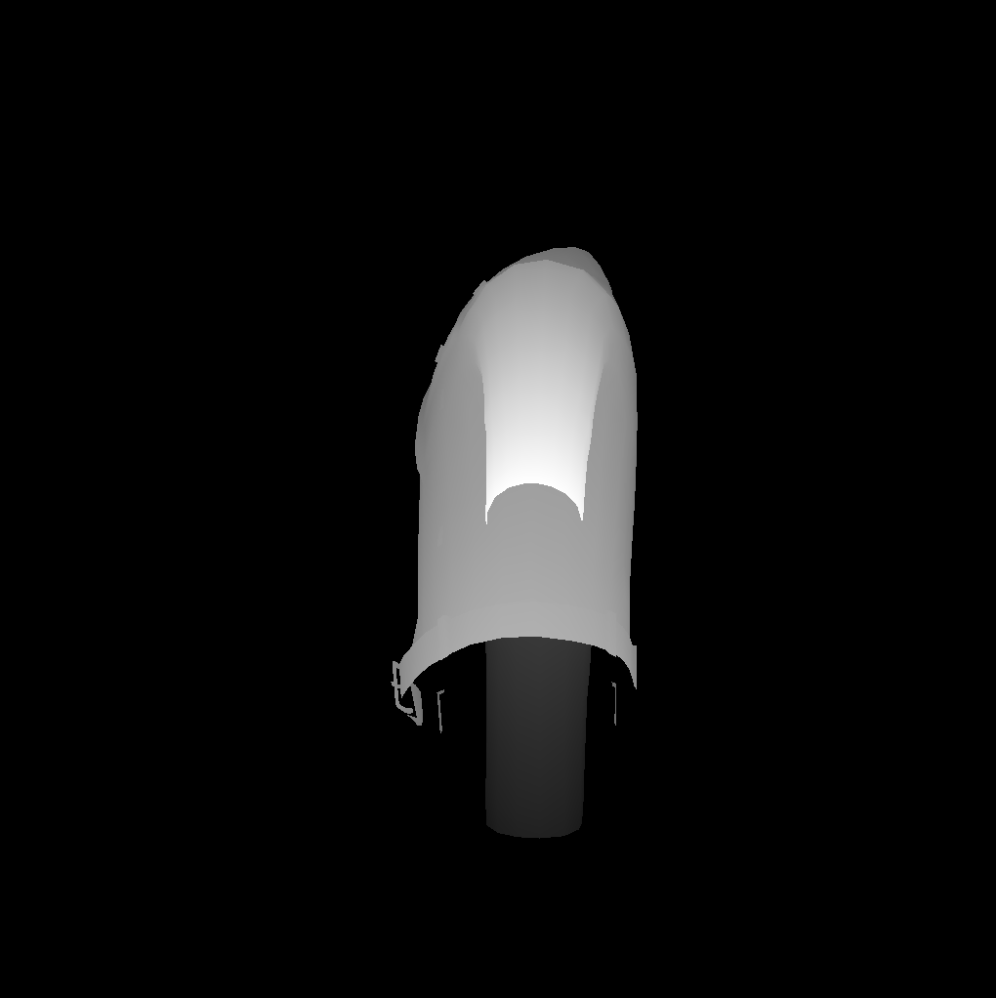}
        \caption{Depth-map}
    \end{subfigure}
    \hfill
    \begin{subfigure}[b]{0.22\linewidth}
        \includegraphics[width=\linewidth]{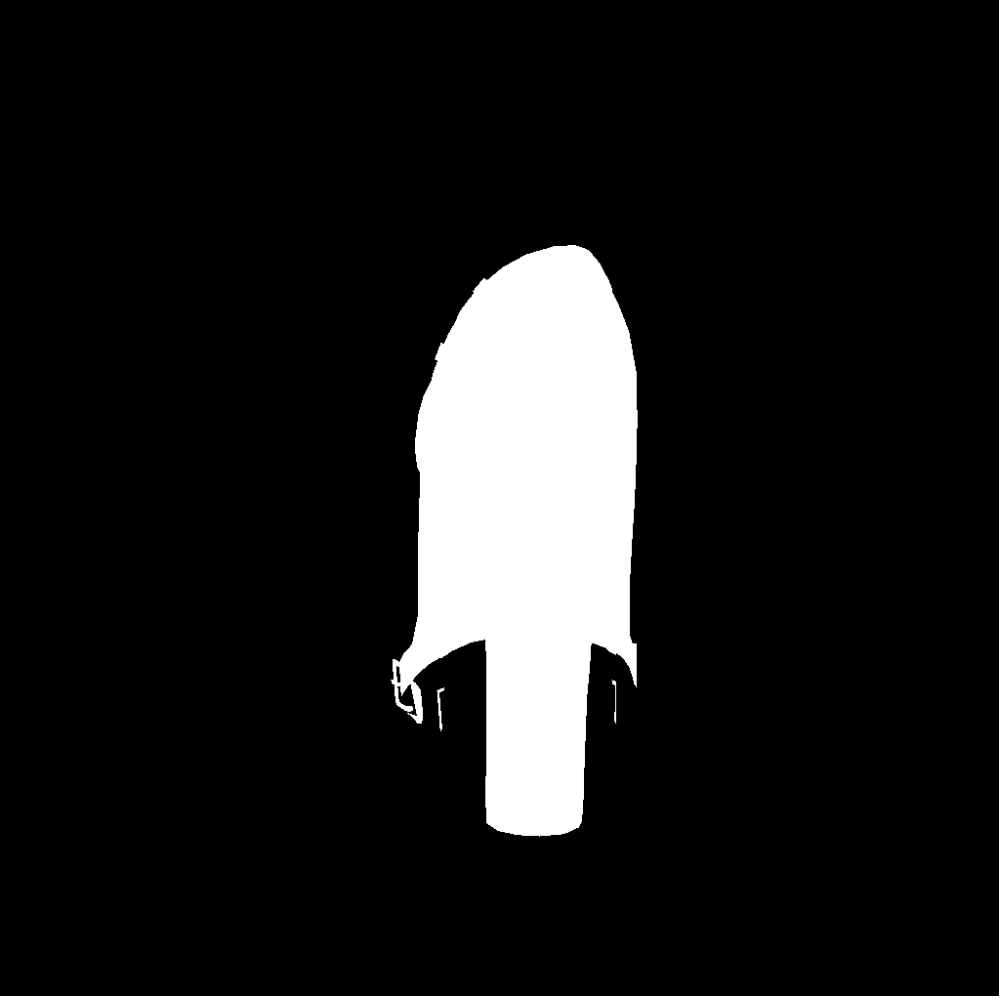}
        \caption{Mask}
    \end{subfigure}
   \hfill
    \begin{subfigure}[b]{0.27\linewidth}
        \includegraphics[width=0.55\linewidth]{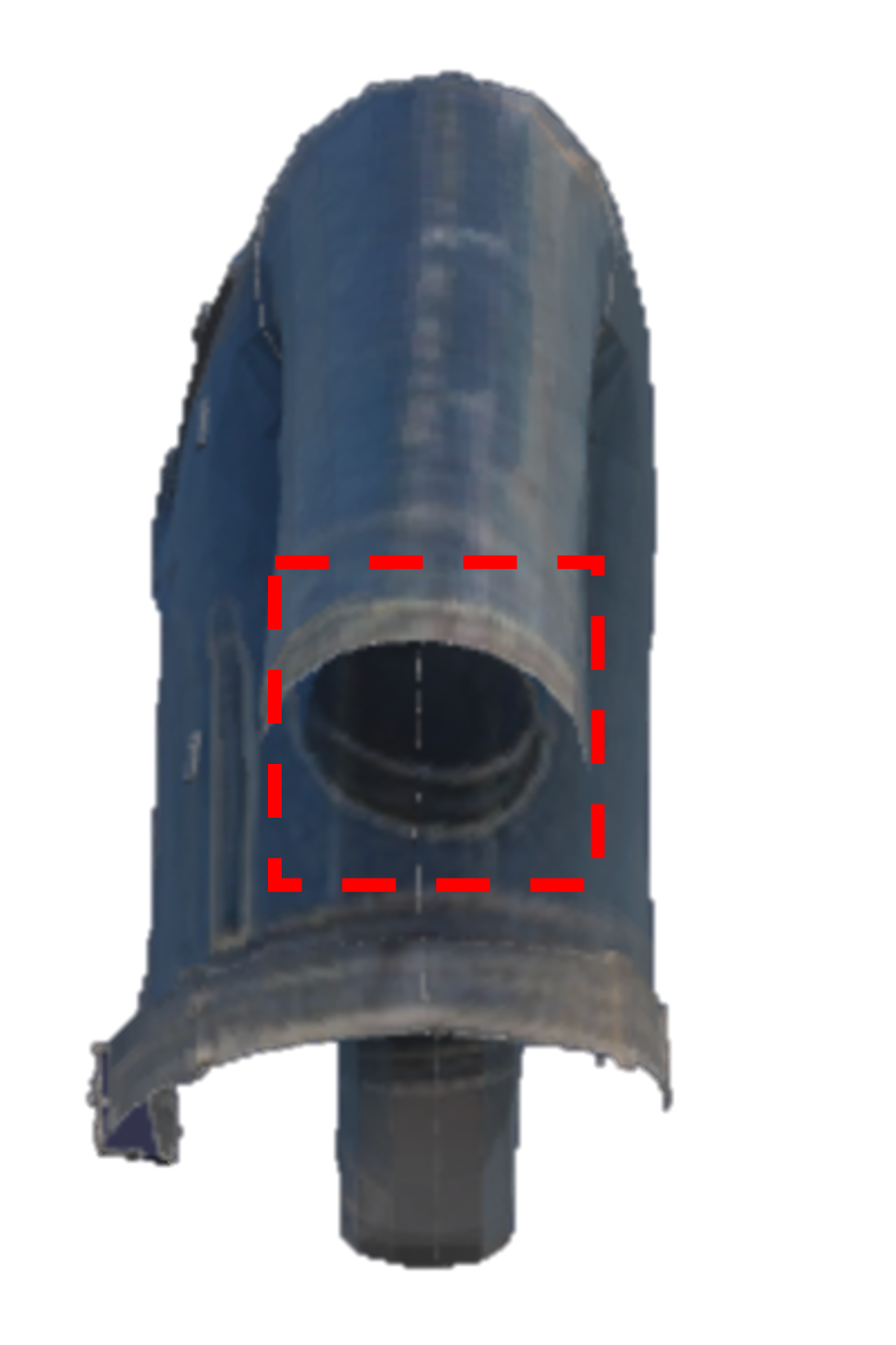}
        \caption{Generated RGB}
    \end{subfigure}
    \begin{subfigure}[b]{0.25\linewidth}
        \includegraphics[width=\linewidth]{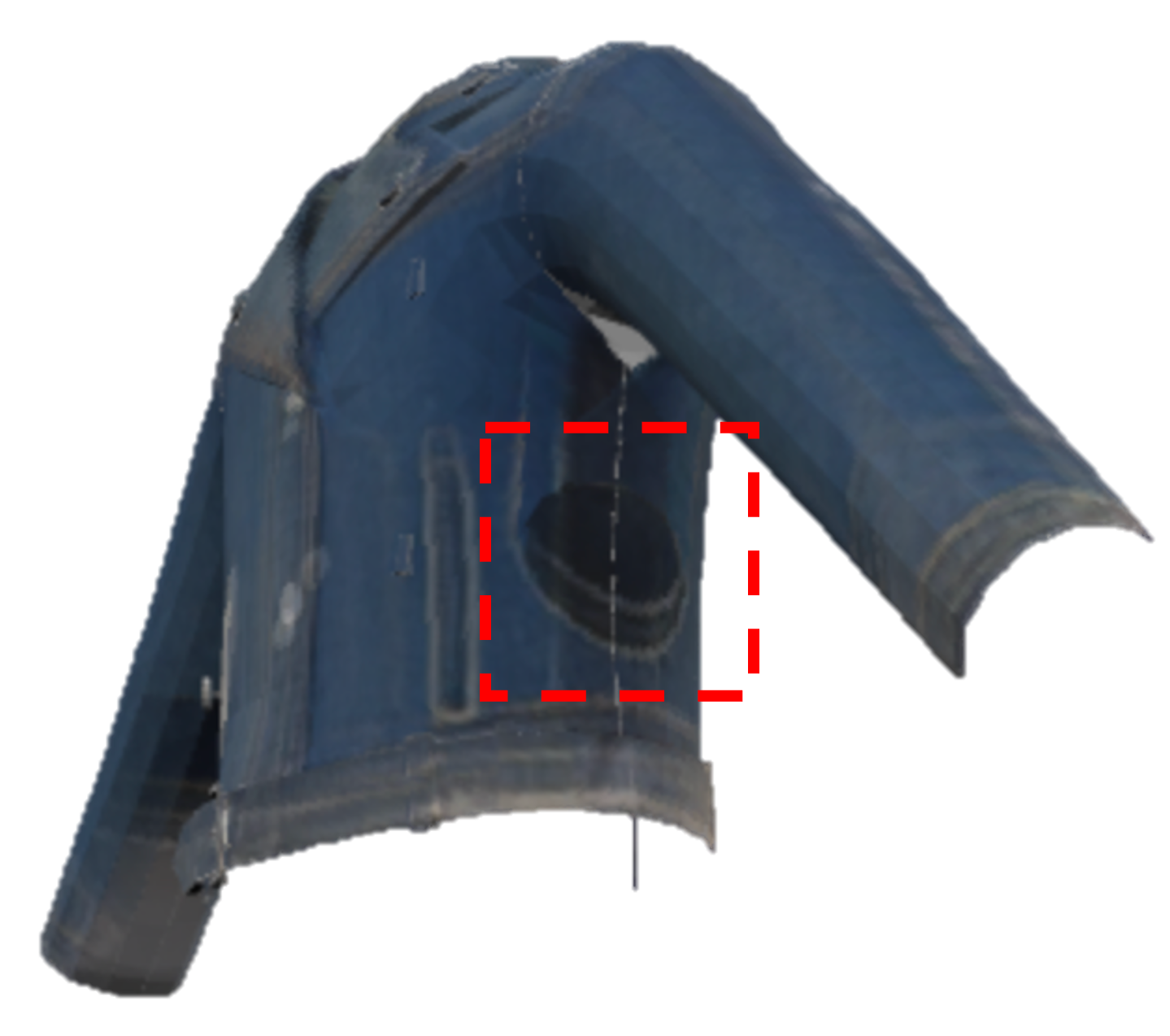}
        \caption{Artifact}
    \end{subfigure}
    \caption{Artifact illustration without rendering internal faces}\label{fig:cull_artifacts}
\end{figure}

To solve this issue, we first utilize backface-culling via the rasterizer to avoid rendering internal faces. However, this alone is insufficient. Depending on the asset and the viewpoint, simply removing non-visible faces can still create new types of artifacts. As shown in Fig. ~\ref{fig:cull_artifacts}, this can create a ``hole" in the generated image, which is then erroneously back-projected onto the the mesh. Unfortunately, even using a rendered mask does not resolve this issue. 




\begin{figure}[tbp]
    \centering
    \begin{subfigure}[b]{0.24\linewidth}
        \includegraphics[width=\linewidth]{imgs/artifact_depth.png}
        \caption{Depth w/ cull-backface}
        \label{fig:depth_w_cullback}
    \end{subfigure}
    \hfill
    \begin{subfigure}[b]{0.24\linewidth}
        \includegraphics[width=\linewidth]{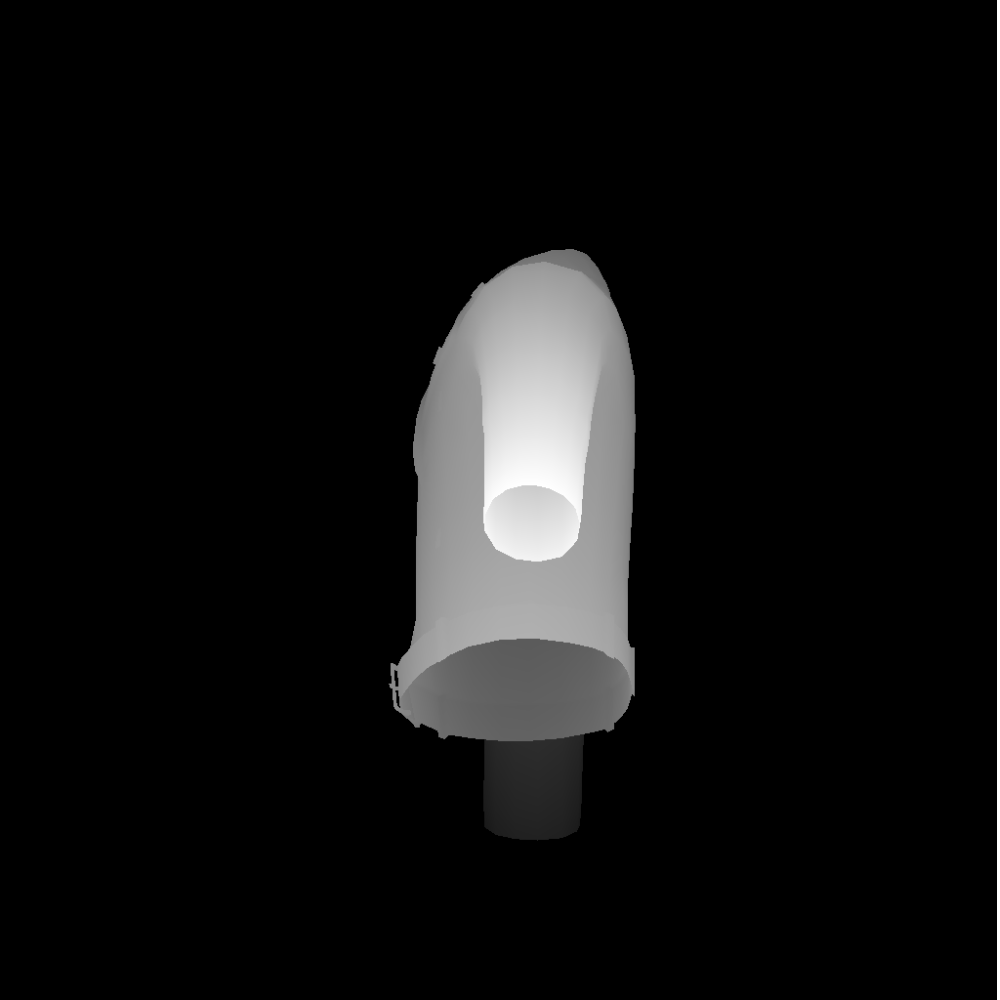}
        \caption{Depth w/o cull-backface}
        \label{fig:depth_wo_cullback}
    \end{subfigure}
   \hfill
    \begin{subfigure}[b]{0.24\linewidth}
        \includegraphics[width=\linewidth]{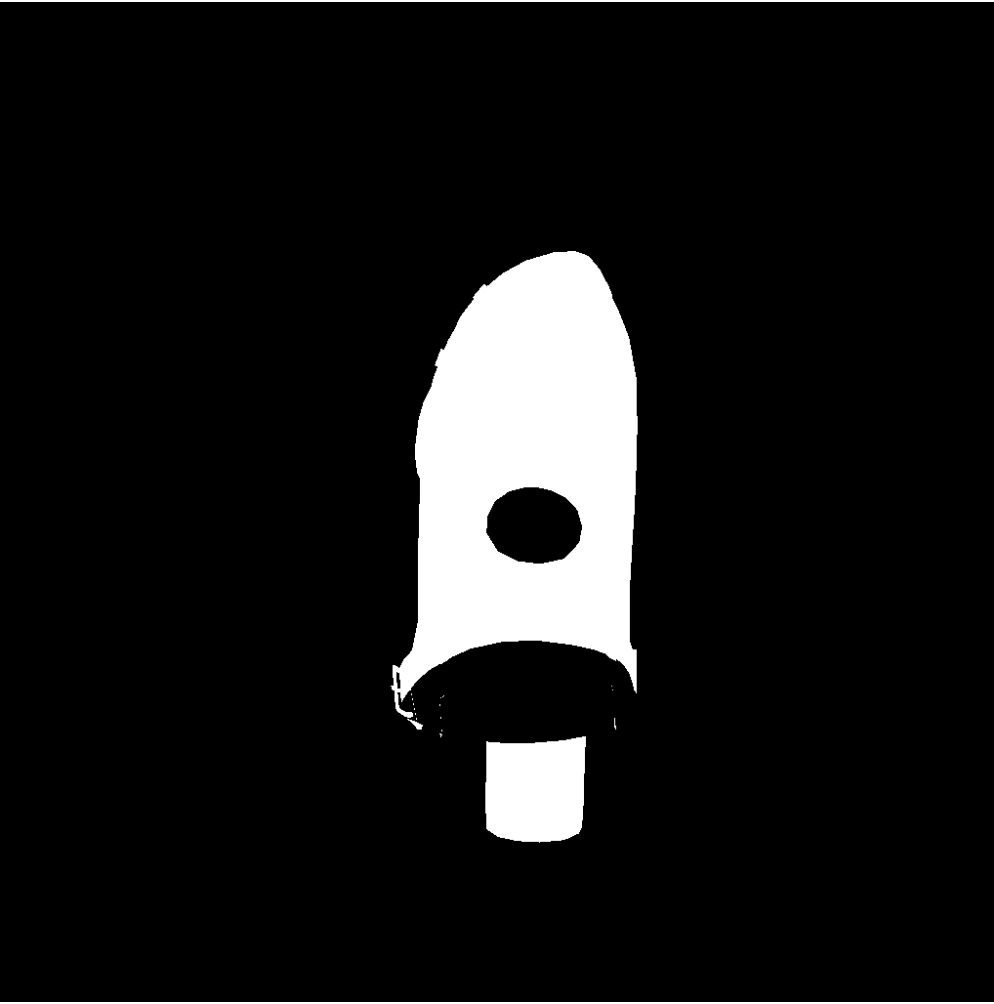}
        \caption{New mask for backprojection}
        \label{fig:new_mask}
    \end{subfigure}
    \begin{subfigure}[b]{0.24\linewidth}
        \includegraphics[width=\linewidth]{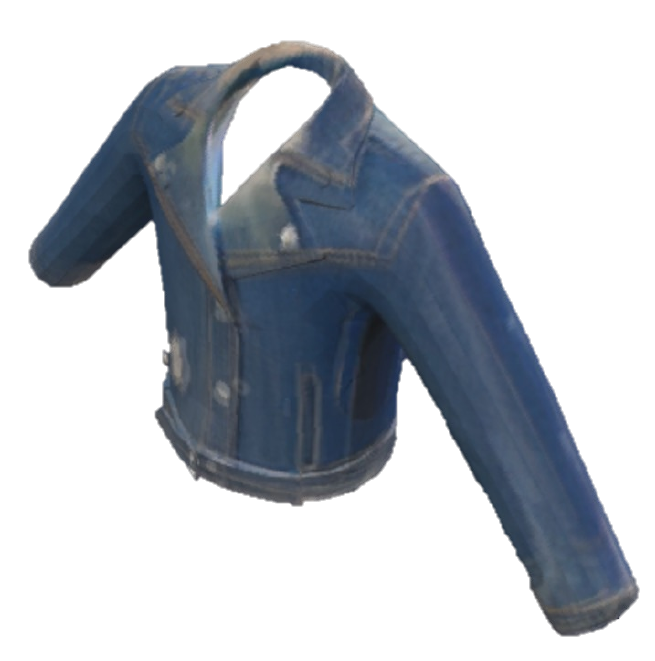}
        \caption{Generation without artifacts}
        \label{new_gen_rgb}
    \end{subfigure}
    \caption{With the new mask computed from the differences of (a) and (b), the artifacts from the internal faces can be avoided in backprojection.}
    \label{fig:open_asset_solution}
\end{figure}

Therefore, we design an improved masking method that ensures only the external faces are backprojected. As illustrated in Fig.~\ref{fig:open_asset_solution}, our solution involves comparing the rendered depths with and without backface culling. This difference gives a new mask, shown in Fig.~\ref{fig:new_mask}, which identifies exactly the internal regions. During the backprojection, the masked pixels from internal faces are not mapped to the texture.




\subsection{Improving Inference Speed}
\label{subsec:imp_speed}

Our texture generation pipeline includes several procedures: text-to-image generation with diffusion models, backprojection and post-processing. Each step is simplified and optimized to improve the inference speed.

\noindent \textbf{Fast Backprojection}
Our fast backprojection technique is designed to efficiently update 3D mesh's UV map upon generated 2D images, while keeping high-quality results. We create the coordinate mapping between UV map and 2D images by rendering the images with 2-channel UV map filled with $xy$-coordinate indices using a differentiable renderer $\mathcal{R}$. The pixels from 2D images are bilinearly splatted to the UV map following the coordinate mapping. The bilinear splatting can be parallel on GPUs to maximize efficiency and minimize aliasing artifact for superior quality.

\noindent \textbf{Reducing Diffusion Model Latency}
The inference time of diffusion models is typically proportional to the number of steps involved. To minimize the required steps, we changed the solver from DDIM~\cite{song2020denoising} to DPMSolver++~\cite{lu2022dpm}, a fast high-order solver for diffusion ODEs with convergence order guarantee. Empirically,  we found that sampling with 20 steps yields high-quality results, and even 10 steps can produce satisfactory outcomes. As shown in Tab.~\ref{tab:latency_red}, changing solver and reducing the steps to 10 decreases the inference time from 19.99 to 12.12 seconds.

We further accelerate the model with torch compile, which JIT-compiles the PyTorch codes into optimized kernels. This compilation was applied to both the UNet-based latent diffusion model and the inpainting and depth ControlNets, reducing the latency further to 10.20 seconds.

As discussed in Sec.~\ref{subsec: geo-aware-mv-gen}, the front and back views are generated simultaneously in grids to ensure the global consistency. We further extend this strategy to all the non-overlapping views, enabling parallel the text-to-image synthesis. This approach reduces the end-to-end runtime to 9.34s.

\begin{table*}[tbp]
\centering
\caption{Latency Reduction. We choose a 6-view texturing process as baseline and compare the influence of each optimization on the end-to-end inference time. Measured on NVIDIA A100, the latency improves from 19.99s to 4.56s.}
\label{tab:latency_red}
\begin{tabular}{c|c|cc}
\hline
\multicolumn{2}{c|}{Optimization}                                  & Time cost (s) & Speedup \\
\hline
\multicolumn{2}{c|}{Baseline (6 views)}                            & 19.99         & 1$\times$      \\
\hline
\multirow{4}{*}{Diffusion model} & +Reduce inference steps        & 12.49         & 1.60$\times$   \\
                                 & +More efficient solver         & 12.12         & 1.64$\times$   \\
                                 & +PT2.0 compile                 & 10.20         & 1.96$\times$   \\
                                 & +Parallel inference            & 9.34          & 2.14$\times$   \\
\hline
Implementation                  & +Rewrite all operations to GPU & 9.23          & 2.17$\times$   \\
\hline
Backprojection Efficiency        & +Fast backprojection                 & 4.56          & 4.38$\times$   \\
\hline
Hardware                        & A100 $\rightarrow$ H100              & \textbf{3.07}          & \textbf{6.51$\times$}   \\
\hline
\end{tabular}
\end{table*}

As shown in Tab.~\ref{tab:latency_red}, the acceleration of diffusion model brings $2.14\times$ speedup. Along with our fast backprojection implementation, our method achieves $4.38 \times$ speedup with a latency of 4.56 seconds on one A100 GPU.
Due to GPU scarcity we developed these improvements on A100, and on H100 we achieve a further reduction to 3.07 seconds.

\section{Experiments}
\subsection{Experimental Setup}

\noindent \textbf{Implementation Details.} Our whole pipeline is implemented in PyTorch~\cite{paszke2017automatic}, with renderers from PyTorch3D~\cite{ravi2020accelerating}. The texture is represented as a $1024 \times 1024$ resolution image. During the texturing stage, we sample 6 views with FoV orthographic camera. Each viewpoint is rendered at $1024\times1024$ size, subsequently cropping the object region and resizing it to $512\times512$ for the diffusion model.

\noindent \textbf{Dataset.} We evaluate the results on 32 objects of diverse categories, including both artist-created and generated meshes~\cite{tang2024mvdiffusion++,sarafianos2024garment3dgen}, with watertight and non-watertight assets.

\noindent \textbf{Baselines.} We compared with the other diffusion-based text-to-texture generation methods: TEXTure~\cite{richardson2023texture}, Text2Tex~\cite{chen2023text2tex}, SyncMVD~\cite{liu2023text}, InTeX\cite{tang2024intex} and Paint3D~\cite{zeng2024paint3d}. We provide the same prompt to each method and render 20 views from each textured mesh for evaluation.

\noindent \textbf{Evaluation Metrics.} We take the following commonly used metrics to evaluate the quality of generated textures: Frechet Inception Distance (FID)~\cite{heusel2017gans}, and Kernel Inception Distance (KID $\times 10^{-3}$)~\cite{binkowski2018demystifying} using clean-fid~\cite{parmar2021cleanfid}. For the user study, we ask an Amazon Mechanical Turk labeler to compare one video with 360-degree rotational view produced by \textit{Make-A-Texture} and one produced by an other method.
We ask labelers two questions: ``{\bf text alignment:} Which object better matches the caption?" and ``{\bf quality:} Aesthetically, which object is better?"
For every data sample, we ask 3 labelers and average the results. Additionally, we compare the runtime on Nvidia A100 GPU.

\subsection{Quantitative Results}

Table~\ref{tab:quant-results} shows that our work outperforms 4 previous methods in FID, KID$\times 10^{-3}$, and the user study. 
While our method underperforms SyncMVD on FID and KID, it ties SyncMVD in the user study.
With SyncMVD running 17$\times$ slower than our method on A100, we approach the SOTA quality with real-time interactive latency. 


\begin{table}[tbp]
\centering
\caption{Quantitative comparison with previous methods (\textcolor{red}{best results} and \textcolor{blue}{2nd-best} results are highlighted in \textcolor{red}{red} and \textcolor{blue}{blue}, respectively). For the user study, a higher number indicates users prefer our method over the prior work, e.g. users prefer the quality of \textit{Make-A-Texture} over TEXTure 71\% of the time.}
\label{tab:quant-results}
\resizebox{\columnwidth}{!}{
\footnotesize
\begin{tabular}{c|cc|cc|c}
\hline
& \multicolumn{2}{c|}{Automatic} & \multicolumn{2}{c|}{User Study Win Rate} & A100 \\ 
Method    & FID $\downarrow$ & KID $\downarrow$  & Quality & Alignment & Runtime (s) \\ 
\hline
TEXTure~\cite{richardson2023texture}   &      123.63            &           27.07  & 71\% & 57\%                 & 90         \\
Text2Tex~\cite{chen2023text2tex}  &       120.37           &   25.11   & 72\% & 64\%                        & 287         \\
SyncMVD~\cite{liu2023text}   &      \textcolor{red}{110.99}            &   \textcolor{red}{20.49} & 51\% & 49\%                         & 81         \\
Paint3D~\cite{zeng2024paint3d}   &    117.39              &  23.67   & 69\% & 66\%                        & \textcolor{blue}{66}         \\
InTex~\cite{tang2024intex}   &      117.07            &     21.25          &    76\%        &   69\%  &  20      \\
\textbf{Ours}      &         \textcolor{blue}{113.99}         &    \textcolor{blue}{20.95}   & - & -                      & \textcolor{red}{4.56}        \\
\hline
\end{tabular}
}
\end{table}

\subsection{Qualitative Results}
\begin{figure*}
	\centering
	\includegraphics[width=0.95\textwidth]{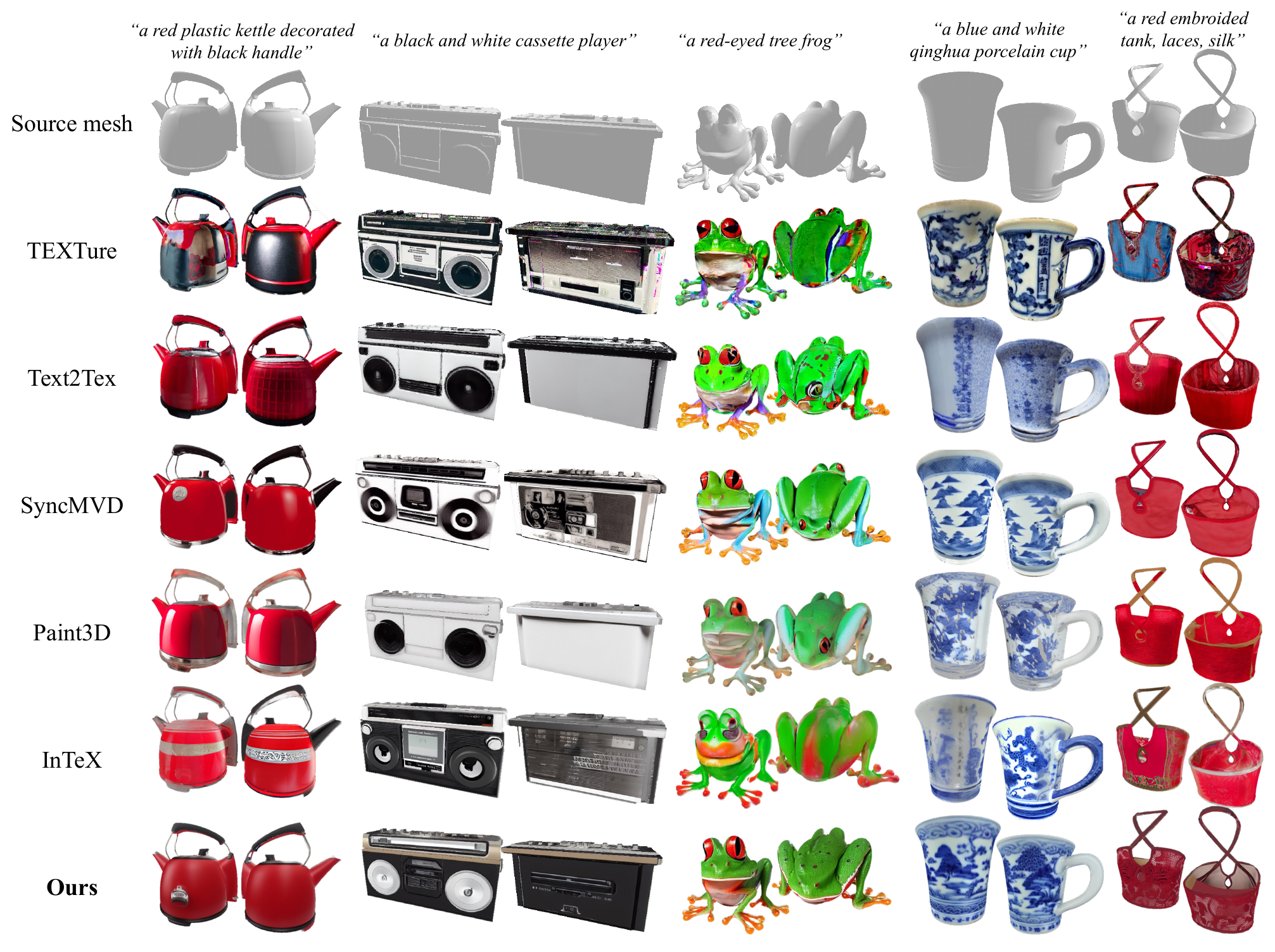}
	\caption{Qualitative comparison with previous works. }
	\label{fig:comp}
\end{figure*}

In Figure~\ref{fig:comp}, we show a visual comparison of our output with those from 4 other previous methods across 5 samples from our evaluation set.
Early methods such as TEXTure and Text2tex struggle with multi-view consistency, \textit{e.g.} the front and back of the teapots are in completely different styles.
More recent methods like SyncMVD and Paint3D produce better multi-view consistency, but struggle with long runtimes of over one minute.
Our \textit{Make-A-Texture} produces results of high visual quality comparable to SyncMVD and Paint3D while achieving latency of just 4.56 seconds on A100 and 3.07 seconds on H100.

\section{Applications}
\label{sec:application}

\subsection{Text-guided Texture Enhancement}
\label{subsec: tex_enhance}
\begin{figure}
	\centering
	\includegraphics[width=\linewidth]{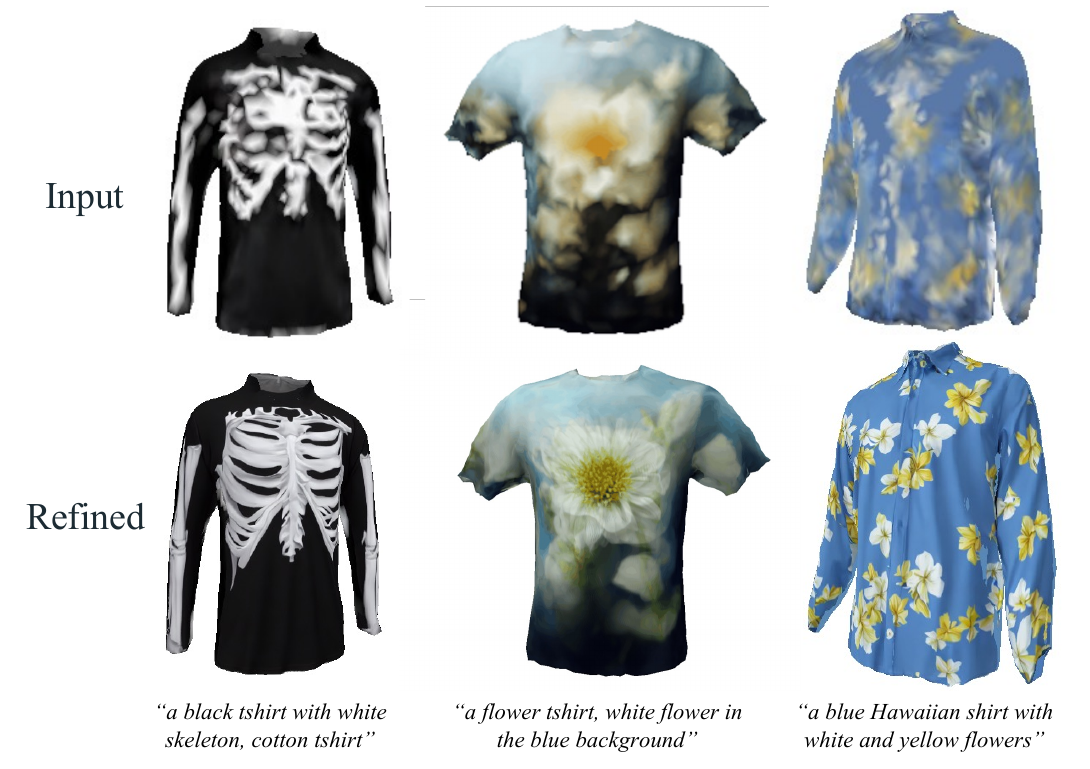}
	\caption{Texture refinement results. Top row shows the meshes generated from~\cite{long2024wonder3d}, and bottom row shows the output refined by our pipeline. By controlling the noise scale, we are able to control the similarity between the input and output.}
	\label{fig:texture_refine}
\end{figure}

Our proposed texture generation pipeline can also be applied to texture editing and enhancement: given a low-quality, low-resolution initial texture $T^{LQ}$, we could extract text prompt from it with a captioning model, and apply our pipeline to refine the details. We utilize SDEdit~\cite{meng2021sdedit}, which perturbs the diffusion sampling process with Gaussian noise and progressively denoises by simulating the reverse stochastic differential equations. This modification allows us to project the low-quality $I_i^{LQ}$ onto the manifold of realistic image $I_i^{HQ}$. By backprojecting these high-quality images, we acquire a refined texture $T^{HQ}$. Fig.~\ref{fig:texture_refine} shows the refinement results from outputs of a 3D generation method~\cite{long2024wonder3d}. Our method can facilitate texture editing tasks like avatar editing~\cite{xiu2024puzzleavatar}.

\subsection{Interactive Fashion Creation}
Due to its remarkably low latency, our method facilitates real-time interactive fashion creation. One application involves virtual reality (VR): upon receiving user instructions, our system efficiently retrieves untextured fashion assets from an extensive collection. Subsequently, it employs our \textit{Make-A-Texture} to dynamically generate detailed textures for these assets. This capability allows users not only to visualize but also to try on the newly textured garments and interact with them freely in VR. 


\subsection{Themed Asset Generation}
\begin{figure}
	\centering
	\includegraphics[width=\linewidth]{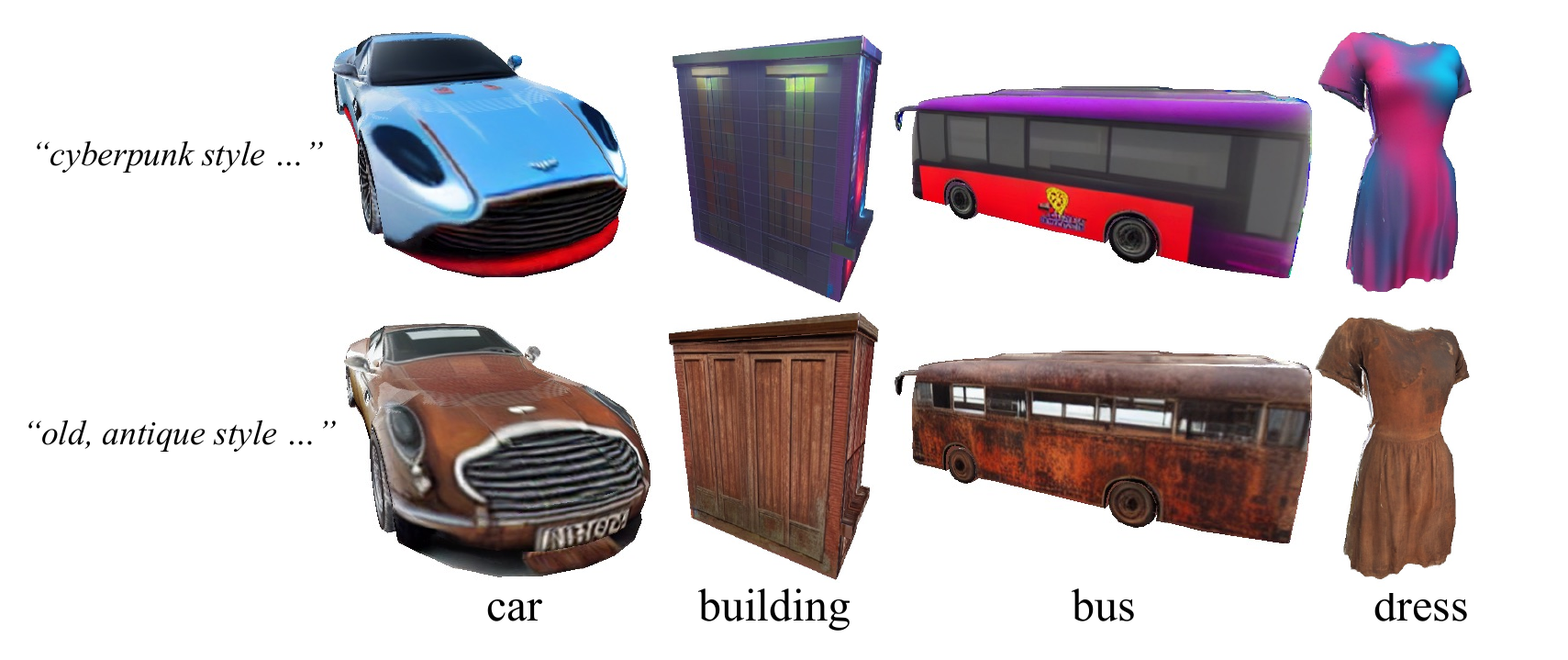}
	\caption{With Make-A-Texture, creators can quickly prototype the design of different objects under one theme.}
	\label{fig:theme_generation}
\end{figure}

\textit{Make-A-Texture} empowers creators to quickly prototype designs across various objects while maintaining a consistent theme, as shown in Fig.~\ref{fig:theme_generation}. Considering the coherence and productivity of designers' workflow, our method is integrated as a Unity plugin~\cite{Unity3D}. With this integration, designers can effortlessly apply visually-coherent textures to multiple objects within the same project.

\section{Ablation Studies}

\subsection{Influence of Filtering Non-frontal Faces}
\begin{figure}[tbp]
	\centering
	\includegraphics[width=\linewidth]{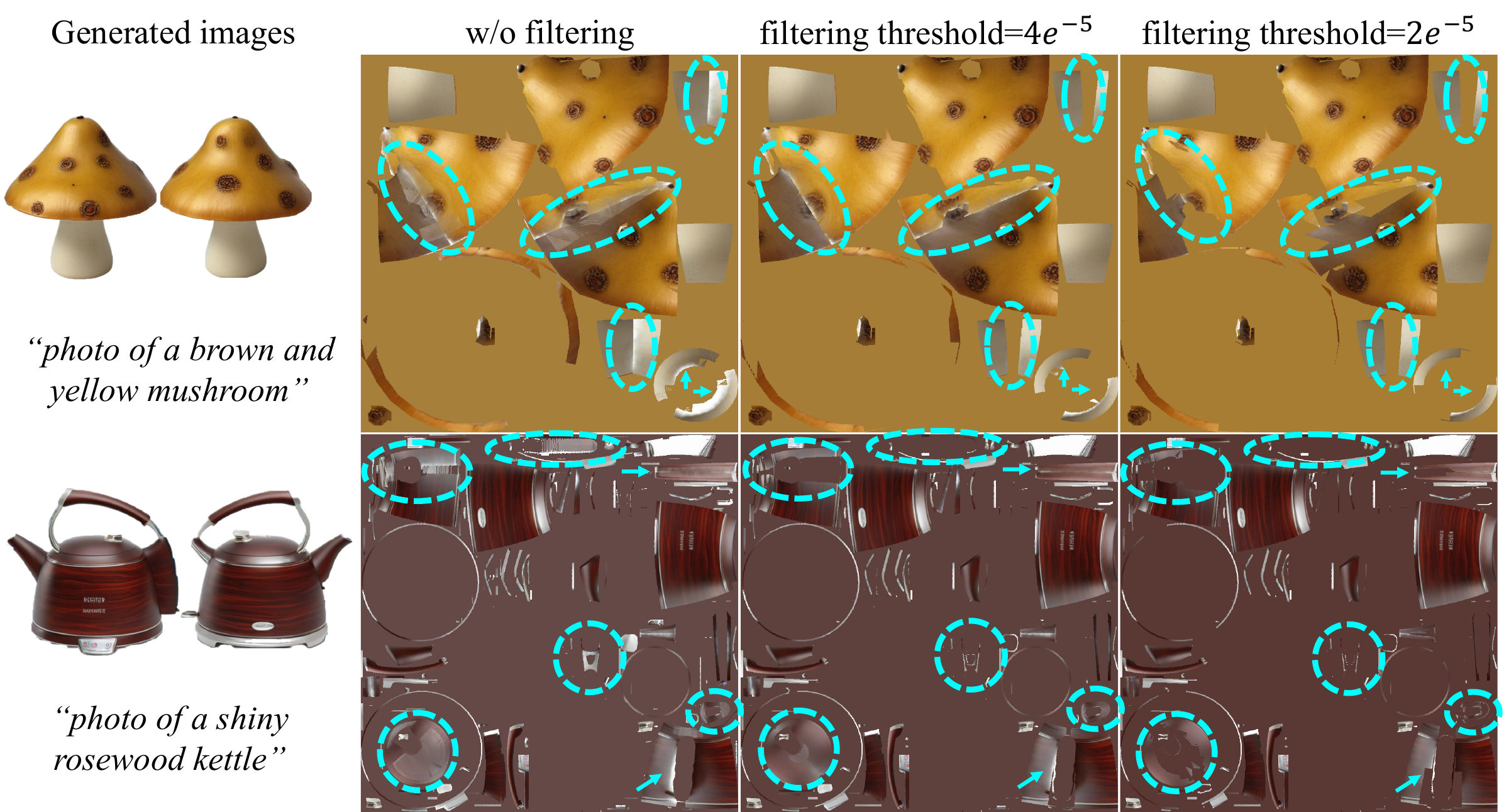}
	\caption{\textbf{Ablation study of filtering non-frontal faces} (zoom in for better visualization). The leftmost column shows the generated front and back view images, the rest columns are backprojected textures. Differences are highlighted in \textcolor{cyan}{cyan}. Filtering out the non-frontal faces can reduce the baked-in highlight and shadows.}
	\label{fig:ablation_filtering}
\end{figure}

Fig~\ref{fig:ablation_filtering} visualizes the influence of filtering non-frontal faces during backprojection. Here we experiment three settings: (1) without filtering, (2) filtering with threshold $4e-5$, and (3) filtering with threshold $2e-5$. By comparing the differences in textures, filtering with lower threshold can mitigates baking the undesirable highlights and shadows, thereby enhancing the overall consistency of the generated textures. 

\subsection{Number of Sampled Views}
We analyze the quantitative results with different number of views during the generation stage in Tab.~\ref{tab:num_of_views}.As the number of viewpoints increases, more textures are painted, leading to an improvement in the FID/KID scores while increasing the runtime. However, these scores begin to plateau or even worsen after more than 6 views.

\begin{table}[tbp]
\centering
\caption{The influence of different number of views. }
\vspace{-2mm}
\label{tab:num_of_views}
\resizebox{\columnwidth}{!}{
\begin{tabular}{ccccc}
\hline
Number of Views & 2   & 4   & 6      & 8      \\
\hline
FID $\downarrow$         & 122.70 & 114.43 & 112.14 & 112.18 \\
KID ($\times 10^{-3}$)$\downarrow$         & 25.15  & 22.28  & 21.34 & 21.29  \\
Runtime (s)     & 1.60   & 2.32 & 3.07   & 3.86  \\
\hline
\end{tabular}
}
\vspace{-4mm}
\end{table}

\subsection{The Order of Sampling Views}
\begin{figure}[tbp]
	\centering
	\includegraphics[width=\linewidth]{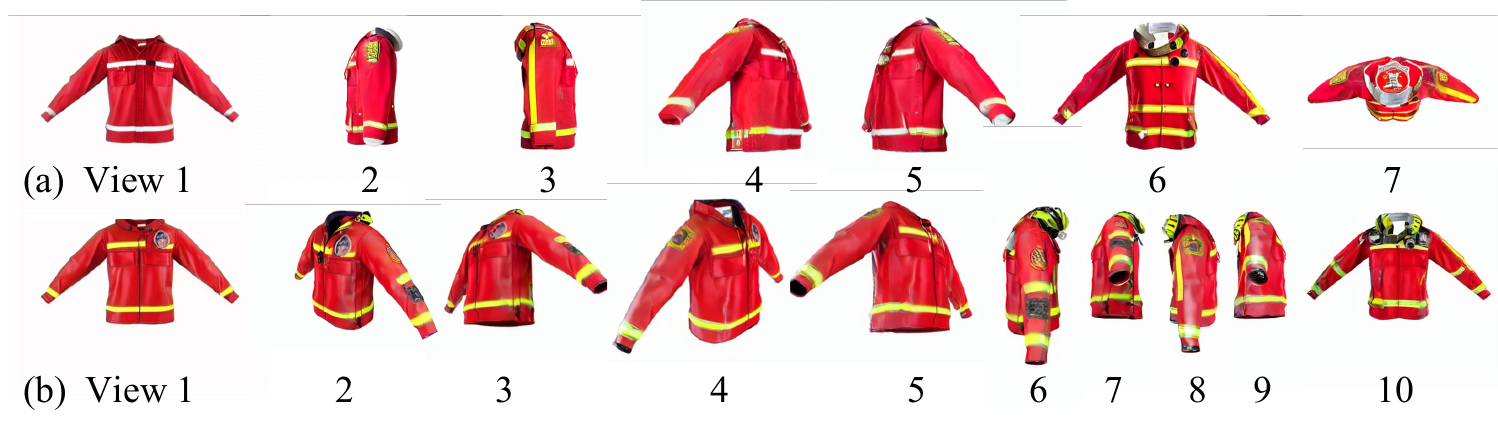}
	\caption{The order of sampling views greatly influence the output. The sampling order of (a) leads to color and pattern shifts in occluded regions like inner sleeve parts, while (b) has more consistent color and patterns. (\textit{Text prompt: a realistic firefighter jacket, highly detailed.)}}
	\label{fig:camera_order}
 \vspace{-3mm}
\end{figure}

Fig.~\ref{fig:camera_order} shows the influence of different orders of sampling views: when rotating the camera, the previous generated textures are only partially visible, thus the new synthesized textures might have a slight color shift due to the partial texture information (see~\ref{fig:camera_order}\textcolor{red}{a}). The inpainting process can only ensure the patterns continuity in the seen views. While for unseen views, the pattern may not be consistent. In \ref{fig:camera_order}\textcolor{red}{b}, the texture consistency is greatly improved with a better sampling and generation order. To evaluate the effectiveness of the automatic view selection algorithm, we compare the performance with a sequence manually labeled by human on the evaluation set. Under the setting of 6 views, the automatic selected sequence achieves comparable FID score (112.14) as the human annotation (112.18).

\section{Limitations}
Our framework still encounters several limitations. Since the foundation model trained on natural images, the lights and shadows are inevitably baked into generated textures. Inpainting sometimes produces disharmonious colors and patterns, affecting the visual coherence of the texture maps. Additionally, the Janus problem is evident, where the back view textures may mirror the front, due to the distribution bias in the depth-aware synthesis training data.
Moreover, the generated textures occasionally do not align with the intended aesthetic style, highlighting the need for improved style control in the generative process, \textit{e.g.} image guidance. Future works will focus on training a multiview generation model to resolve the Janus problem, using rendered training data with only ambient light to remove the lights and shadows, refining inpainting model, and incorporating style control to better meet aesthetic expectations.

\section{Conclusions}
We presented \textbf{Make-A-Texture} framework, an efficient solution for generating high-resolution textures for given 3D geometries using textual prompts. 
By integrating a depth-aware inpainting generation process with an automatic view selection algorithm, our method ensures geometric fidelity and consistency across complex 3D models. 
A key achievement of our method is its speed, capable of producing 1024$\times$1024 texture maps in approximately 3 seconds on a single H100 GPU. This remarkable efficiency opens a door for real-time applications for the first time. 
Additionally, we enhance the output texture quality by handling highlights and shadows, and expand the framework's capability to texture open-surfaced assets. \textit{Make-A-Texture} makes texture generation a more applicable tool for artists and general users in gaming and virtual reality.

{\small
	\bibliographystyle{ieee_fullname}
	\bibliography{main}
}

\end{document}